\documentclass[journal]{IEEEtran}

\usepackage[utf8]{inputenc}
\usepackage{graphicx}
\usepackage{color, soul}
\usepackage{transparent}
\usepackage{import}
\usepackage{rotating}
\usepackage{booktabs}
\usepackage[table,dvipsnames]{xcolor}
\usepackage[english, status=draft]{fixme}
\usepackage[linesnumbered, ruled]{algorithm2e}
\fxusetheme{color}
\usepackage{amsmath}
\usepackage{amssymb}
\usepackage{subfigure}
\usepackage{pifont}
\usepackage{multirow}
\usepackage{lscape}
\usepackage{scrextend}
\usepackage{hyperref}
\usepackage{mathrsfs}
\usepackage{hyperref}

\usepackage{xspace}
\makeatletter
\DeclareRobustCommand\onedot{\futurelet\@let@token\@onedot}
\def\@onedot{\ifx\@let@token.\else.\null\fi\xspace}
\def\eg{\emph{e.g}\onedot} 
\def\ie{\emph{i.e}\onedot}

\makeatother

\begin{document}
\title{The Visual Social Distancing Problem}

\author{Marco Cristani,~\IEEEmembership{Member,~IEEE,}
        {Alessio Del Bue},~\IEEEmembership{Member,~IEEE,}
        {Vittorio Murino},~\IEEEmembership{Senior Member,~IEEE,}
        {Francesco Setti},~\IEEEmembership{Member,~IEEE,}
        and~Alessandro Vinciarelli,~\IEEEmembership{Member,~IEEE}% <-this % stops a space
\thanks{All the authors equally contributed to this manuscript and they are listed by alphabetical order.}% <-this % stops a space
\thanks{M. Cristani is within Humatics SRL (\url{www.humatics.it}) and University of Verona; A. Del Bue is within Pattern Analysis and Computer Vision (PAVIS) research line of the Istituto Italiano di Tecnologia (IIT); V. Murino is within  University of Verona, Italy, and the Ireland Research Centre of Huawei Technologies Co., Ltd.; F. Setti is within the department of Computer Science of the University of Verona, Italy; and A. Vinciarelli is within the School of Computing Science  of the University of Glasgow, UK.}
}

\markboth{ }%
{Cristani, Del Bue, Murino, Setti, Vinciarelli: The Visual Distancing Problem}



\maketitle

\begin{abstract}
One of the main and most effective measures to contain the recent viral outbreak is the maintenance of the so-called Social Distancing (SD). To comply with this constraint, workplaces, public institutions, transports and schools will likely adopt restrictions over the minimum inter-personal distance between people. 
Given this actual scenario, it is crucial to massively measure the compliance to such physical constraint in our life, in order to figure out the reasons of the possible breaks of such distance limitations, and understand if this implies a possible threat given the scene context.
%%Marco mia aggiunta sotto
All of this, complying with privacy policies and making the measurement acceptable.
To this end, we introduce the Visual Social Distancing (VSD) problem, defined as the automatic estimation of the inter-personal distance from an image, and the characterization of the related people aggregations.
%%MARCO: qualcosa non gira nella frase: ho provato a metterla come girava bene a me (togliendo "in"), ma in inglese ho solo da imparare.
% and in the characterization of the related people aggregations.
VSD is  pivotal for a non-invasive analysis to whether people comply with the SD restriction, and to provide statistics about the level of safety of specific areas whenever this constraint is violated. We then discuss how VSD relates with previous literature in Social Signal Processing and indicate which existing Computer Vision methods can be used to manage %fix %solve 
such problem. 
%%MARCO: "fix" dovrebbe indicare più un problema molto specifico da risolvere, un guasto. Provo con manage
We conclude with future challenges related to the effectiveness of VSD systems, ethical implications and future application scenarios. 
\end{abstract}


% \fi
\IEEEpeerreviewmaketitle

%\section{Social distancing problem}

%\IEEEPARstart{T}{his} demo file is intended to serve as a ``starter file''
%for IEEE journal papers produced under \LaTeX\ using
%IEEEtran.cls version 1.8b and later.
%I wish you the best of success.

 

%
%%%% PEZZO DI ALESSANDRO
%
\section{Introduction}
\label{sec:introduction}
Humans are  social species as demonstrated by the fact that in everyday life people continuously interact with each other to achieve goals, or simply to exchange states of mind. % Moreover, as a result of the evolution, the success of our species is also due to our social intellect, to our ability to function as social beings, allowing us to live in groups and share skills and purposes. 
One of the peculiar aspects of our social behavior involves the geometrical disposition of the people during an interplay, and in particular regards the interpersonal distance, which is also heavily dependent on cultural differences. 
%It is so intrinsically embedded and natural in humans that we do not pay much attention to it until limitations are imposed for whatever reasons, 
However, the recent pandemic emergency has affected exactly these aspects, as the extraordinary capability of COVID-19 coronavirus of transferring between humans has imposed a sharp and sudden change to the way we approach each other, as well as rigid constraints on our inter-personal distance. 

This recently imposed restriction is widely, but imprecisely, referred to as ``\emph{social distancing}'' (SD) since prevention of the virus diffusion does not require us to weaken our social bonds. 
%\vm{la frase sopra non gira bene}
The likely reason of SD naming is that, from a cognitive point of view, physical and social aspects of distance are deeply intertwined~\cite{kendon1990conducting}, a phenomenon that popular wisdom captures through a proverb that, in slightly different versions, appears in different languages and cultures, namely ``\emph{far from eyes, far from heart}''. 

Not surprisingly, the time spent in physical proximity with others, in opposition to the time spent in individual activities, is a crucial factor in the ``\emph{social brain hypothesis}'', one of the most successful theories of human evolution~\cite{Dunbar2014}. Similarly, \emph{Attachment Theory}, probably the development model most widely accepted in child psychiatry, revolves around the ability of children and parents to establish and maintain physical proximity~\cite{Bowlby1969}. Finally, the different modulation of interpersonal distances is  known to be one of the main obstacles in intercultural communication~\cite{Hall1959}. 

\begin{figure}
    \centering
    \includegraphics[width=\columnwidth]{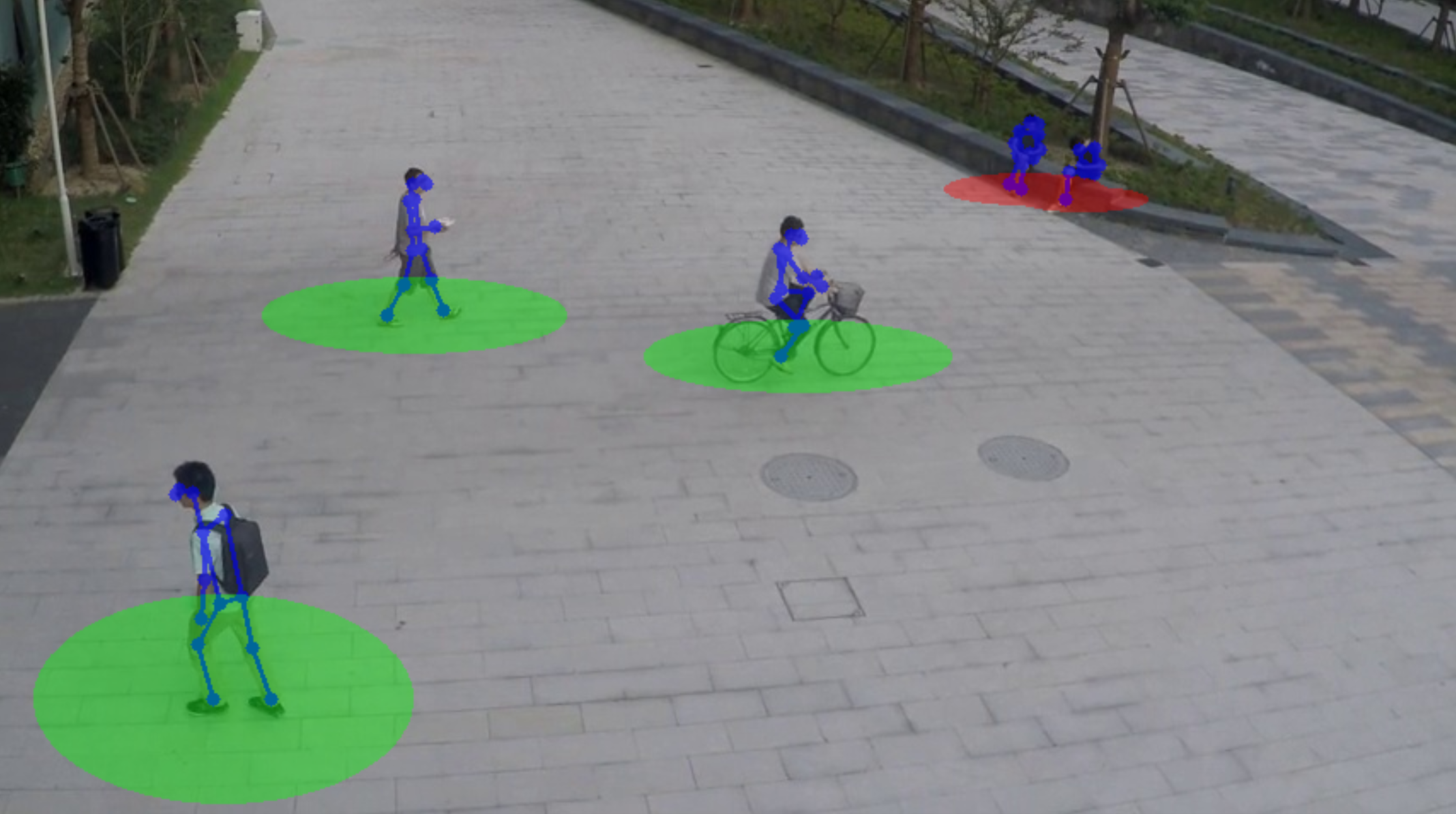}
    \caption{The VSD can be estimated in a single frame as the interpersonal distance between people (a 1m radius disk in this example). People that are closer than the imposed distance (red disks), \ie not respecting geometry, might still respect public rules if having a bond of kinship. Results obtained using the following code: \url{https://github.com/IIT-PAVIS/Social-Distancing}.}    \label{fig:social_dist_ex}
\end{figure}

The above suggests that dealing with interpersonal distances means to deal with evolutionary, developmental and cultural forces that shape, to a significant extent, our everyday life. As a consequence, the role of technologies for the analysis of such distances becomes crucial during pandemics, given that they must mediate between the forces above, responsible for the human tendency to get too close to avoid contagion, and the pressure of prophylactic measures, artificially designed to fight a pathogen inaccessible to our senses and  cognition. 

\begin{figure}
    \centering
    \includegraphics[width=\columnwidth]{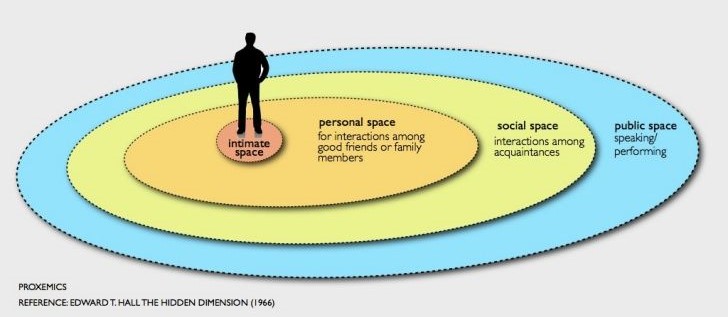}
    \caption{A graphical representation of the personal spaces that are used in proxemics.}
    \label{fig:pers_space}
\end{figure}

% AV: Forse possiamo spezzare il paragraph qui.
One possible solution is to go beyond simply measuring how far we are are from one another, 
%%Marco: mi attacco a quello che abbiamo scritto
as most of the applications on the market are doing (see Sec.~\ref{sec:sd_chara})
and try to make sense of what distances mean. In other words, to inform technologies with principles and laws of \textit{Proxemics}, 
%%Marco: ho messo il maiuscolo, perchè è la regola che abbiamo adottato sin qui.
the psychology area showing how people convey social meaning through interpersonal distances and, ultimately, how social and physical dimensions of space interplay with one another~\cite{Richmond1999}. 
Proxemics is strictly linked to the definition of people gatherings, namely groups, and as such, it depends on its spatial organization and the number of people involved. %In general, the surrounding space around a person is characterized by interpersonal distance \cite{hall1966hidden}, namely: intimate, personal, peri-personal or social, and public spaces (see Fig.~\ref{fig:pers_space}), all associated to different SDs, in turn, also dependent by the degree of kinship and familiarity between the subjects and by the geometrical configuration and size of the environment in which an interplay occurs.
%%MARCO: Ho dei dubbi sul periodo sopra. Le 4 distanze le citiamo qui, ma a che senso? non le usiamo piu' da nessuna parte, e non hanno nessuna interazione con la social distance, e nemmeno con i metodi usati. in pratica, ora come ora mi sembrano appese. Un aggancio potrebbe essere quello di generare una frase ad effetto come quella che i metodi di social distancing, se applicati alla lettera, potrebbero eliminare la presenza di una di queste classi di distanza! Provo a scriverlo qui sotto, rivedendo l'intero periodo.
In general, the surrounding space around a person is characterized by \emph{interpersonal distance classes} \cite{hall1966hidden}, namely: intimate, personal, peri-personal or social, and public spaces (see Fig.~\ref{fig:pers_space}), all associated to different SDs, in turn, also dependent by the degree of kinship and familiarity between the subjects and by the geometrical configuration and size of the environment in which an interplay occurs. A blind application of social distancing rules, encouraging to stay further than 1-2 meters, will eliminate an entire interpersonal distance class and all of the social interactions which take play within it, including for example those between children and relatives.
As can be noticed, behavior, social interactions, and space arrangements are tightly coupled, and affect each other. This is why it is important to take into considerations all these aspects when constraints in this respect are to be imposed, in particular when people health is in play.

For all these reasons, the focus of this paper lies on \emph{Visual Social Distancing} (VSD), \ie on approaches relying on video cameras and other imaging sensors (see Fig.~\ref{fig:social_dist_ex} for an example) to analyse the proxemic behaviour of people.
% AV: forse possiamo evitare di spezzare il paragraph qui.
%%Marco: anche per me, levo.
The main reason behind the choice of VSD is that computer vision and social signal processing have already developed methods for automatic measurement and understanding of interpersonal distances  (see Sec.~\ref{vsdext} for more details). Furthermore, VSD approaches have shown advantages that can complement other technologies like, \eg, mobile applications based on large-scale mobility patterns. In particular, VSD approaches can characterize interpersonal distances in terms of social relations (\eg, whether people at a certain distance are friends, family members or partners)%~\cite{XXX}
, thus allowing one to modulate interventions according to such an information.
%%MARCO si deve ompletare qui, in base allo stato dell'arte di Seti che devo ancora leggere
Furthermore, vision-based technologies can detect contextual information helpful to understand whether social distancing rules are actually being broken or not. For example, VSD can understand whether people get too close because the situation makes it necessary (\eg, when someone rescues a person in troubles) or whether the distance is not a problem (\eg, when people wear personal protective equipment and can safely stay close).
%%MARCO: anche qui sopra bisogna suddividere le referenze riportate immediatamente più sopra
%%MARCO: aggiungo anche un pezzo discusso nel seguito, sul fatto che VSD indica anche le ragioni del perché si ha infringement della SD permessa
Finally, VSD helps to understand the reason why some people stand close, distinguishing whether they are socializing among themselves, or if they are interacting with the environment (as for example looking at a timetable in the airport) thus suggesting the most proper countermeasure to ensure SD (eg. rising an audio alarm to discourage social interactions or putting markers into the floor so that people can watch the time table while keeping the right distances). 

The advantages above appear to be of particular importance since at the moment social distancing rules have to be expressed in simplistic terms (\eg, people have to be at least 2 meters far from one another) that require one to distinguish between the intention (avoid contagion) and the rule (keep a minimum interpersonal distance). Such a distinction, evident to humans, poses a real and new challenge to a computational algorithm for VSD that could solve the problem by leveraging, for instance, the use of contextual information. Differently, the number of false alarms would be so high that any benefit resulting from the use of technology would be canceled.

In the following, we will discuss in detail the VSD problem and its connection to the Computer Vision and Social Signal Processing research domains. Starting from a geometrical point of view, \ie estimating inter-personal distances between people from an image, we show that this first step does not take into account the scene and social contextual. For this reason, a further stage needs to elaborate the geometrical VSD in order to interpret if the violation of the distance is a real cause of alert or an acceptable situation (\eg a family walking together). We then contextualise the VSD in a range of application that can benefit from its application and finally conclude with a description of the possible ethical shortcomings of the application.    
%%%%% FINE PEZZO DI ALESSANDRO

\section{Visual Social distance estimation}\label{vsdext}

%\mc{\textbf{[Alessio]}: how to find distances among people: scenarios (\ie rgb camera, depth, sensor networks etc.) and algorithm}

\begin{figure}
    \centering
    \includegraphics[width=\columnwidth]{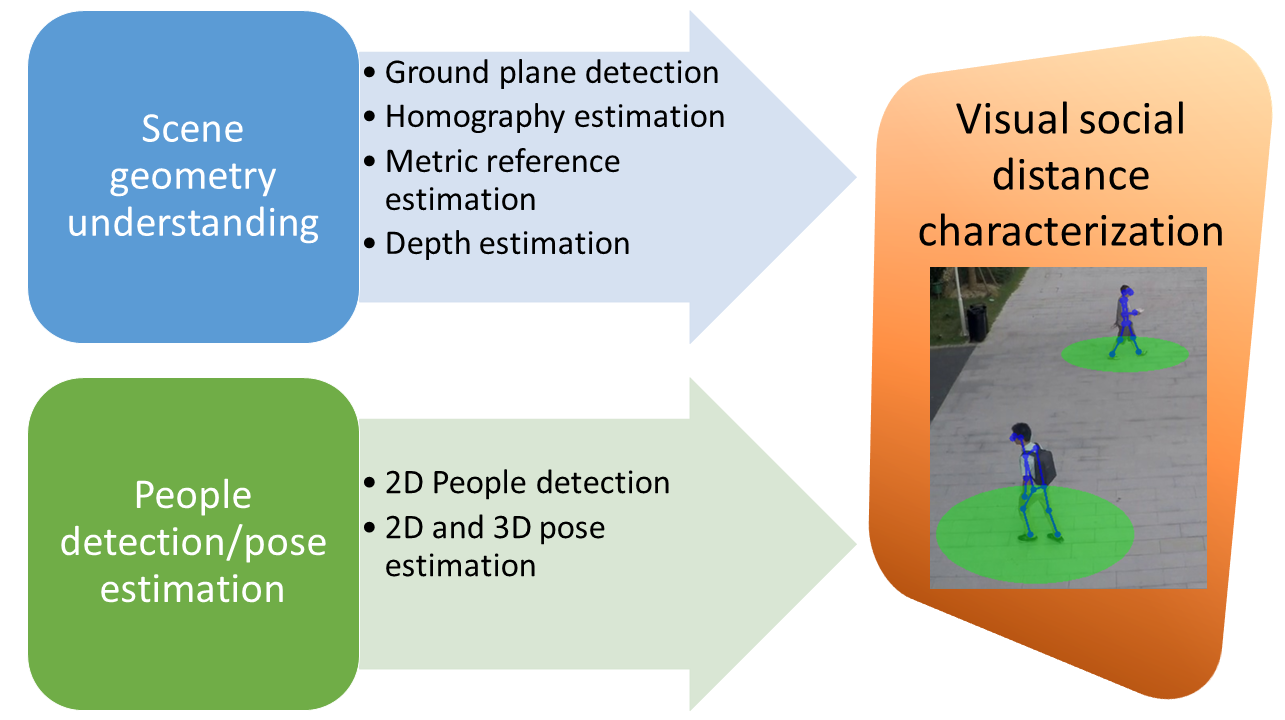}
    \caption{The VSD problem requires the solution of different problems. The estimation of a local metric reference system using the scene geometry (blue box) and the detection of pedestrians and (possibly) their pose in the image (green box). This information provides a geometrical measure of interpersonal distance that has to be interpreted given the social context of the scene (orange box).}
    \label{fig:social_dist_problem}
\end{figure}

Estimating the VSD %has a strong geometrical component that 
requires the solution of a set of classical Computer Vision and Social Signal Processing tasks 
%%MARCO: maiuscolo, ma forse usati come aggettivi vanno lasciati minuscoli
as identified in Fig.~\ref{fig:social_dist_problem}, namely, scene geometry understanding (Sec.~\ref{sec:local_geometry}), person detection/body pose estimation (Sec.~\ref{sec:person_detection}) and social distance characterization (Sec.~\ref{sec:sd_chara}).
%%Marco mettiamoci d'accordo per usare Fig. e non Figure, la maggior parte ora è così
%
%An initial solution of a geometric problem is necessary to obtain a metric reference among people in a scene that has to be further processed by understanding the social context in the image. 
%
Indeed, %As the final aim is to estimate a metric distance among a group of individuals, 
the geometry of the scene is a first important to define a local reference system for measuring inter-personal distances. Clearly, a second and important task is the detection of people in the scene in possibly crowded environments. Once the target people are correctly localised in a scene, their distance can be locally estimate in order to understand if the mutual distance is lower than a threshold (\eg 1m or 2m). Afterwards, this metric information is analysed to output whether there is a violation of the protocol or the short distance is due to a legitimate situation, \eg a family walking together.
%... \adb{MARCO, FRANZ: please finalise this, what group analysis should out put here...}

In the following, we will describe these modules in detail re-targeting, when possible, previous Computer Vision
%%MARCO: messo maiuscolo
methods that can provide a solution to these problems.

%%MARCO: nella figura non andrebbero messi anche i range delle singole distanze?

\subsection{Scene geometry understanding}
\label{sec:local_geometry}
The task of measuring social distancing from images requires the definition of a (local) metric reference system. This problem  is strongly related to the single view metrology topic \cite{criminisi2000single} as we consider the most common case of a fixed camera. An initial solution for estimating inter-personal distances requires the identification of the ground plane where people walks \cite{lee2000monitoring,se2002ground,hoiem2007recovering,groundnet:2019,Abbas:az:2019,wagner2019,watson-2020-footprints}. Such ground plane serves in many video-surveillance systems to visualise the scene as a bird's eye view for ease of visualisation and data statistics representation. Most works impose the assumption that the ground plane is planar. Then, the problem is to estimate an homography given some reference elements (\eg, known objects or manual measurements) extracted from the scene or using the information of detected vanishing points in the image \cite{magee1984determining, criminisi2000single, rother2002new, mirzaei2011optimal, wildenauer2012robust, bazin2012globally, bazin20123, li2012vanishing, zhang2016vanishing, lu20172}. Another common approach is %to %attempt 
to calibrate fixed cameras by observing the motion of dynamic objects such as pedestrians \cite{liu2011surveillance, tresadern_reid_ivc2008,lv2006camera,tang2019esther}. Recently, approaches based on deep learning attempts at estimating directly camera pose and intrinsic parameters on a single image \cite{hold2018perceptual, Lopez_CVPR_2019}.

Even if these approaches might provide an estimate of the camera intrinsic/extrinsic parameters and the detection of the ground plane, still VSD estimation requires a metric reference. Such information can be coarsely computed in the scene given objects of known dimension or by using a standardised height of pedestrians as a rule of thumb \cite{wagner2019, benabdelkader2008statistical, vester2012estimating}. Given the current state of the art, we have the following observations related to the geometrical aspects of VSD:
\begin{itemize}
    \item Although the planarity constraint might not hold for the entire image, VSD has to do a local estimation of proximity for which is safe to relax the scene being piece-wise planar.
    \item Self-calibration approaches highly rely on the existence of a Manhattan world (\eg vanishing points are detectable) or pedestrian walking in straight trajectories, which limit the applicability of such methods. Depth from single image might be a viable option, but a metric reference is still needed.
    \item Estimating a metric reference for precise SD measures from images is an issue. Such reference extracted from pedestrians might be unreliable given the variations in anthropometric characteristics. Reasoning on the geometrical context  of the scene (\eg,  object shapes) can lead to a more robust metric estimate. 
\end{itemize}

It is also important to emphasizes here that VSD is a simpler problem than estimating every metric distances among people in any position in the image. SD is necessary when two or more pedestrians get close enough for triggering the necessity of a measure. At this point, a local reference system can be estimated and metric references can be leveraged by using surrounding objects and the height of the local cluster of people.
%%MARCO: non si capisce se questa sia una assunzione tua oppure una regola implementata da qualche approccio

%\begin{figure}
%    \centering
%    \includegraphics[width=\columnwidth]{./F-Formation.jpg}
%    \caption{ZZZ. SCEGLIERE UNA TRA QUESTA E QUELLA DOPO}
%    \label{fig:f-form}
%\end{figure}

To this end, the SSP problem of detecting and tracking the formation of groups~\cite{bazzani2013social,Cristani:FF:BMVC:2011,ge2012vision,setti2015plos,solera2016pami} %\adb{(FIXME MARCO/FRANZ REFERENCES)} 
can be useful for selecting which pedestrians should be used as a subset for estimating the local VSD. These local estimates with an associated metric reference can be useful whenever a global camera pose is hard to estimate or if the single ground plane  assumption is violated, a likely occurrence in an unconstrained scenario.     
%%MARCO: anche qui non capisco: dici che per stimare la distanza di due persone ti puoi focalizzare su un sottoinsieme di esse, per esempio quelle che realizzano un gruppo, e construire un sistema di riferimento attorno ad esse e stimare la distanza. Giusto? Mi sembra abbastanza complicato, ci sono riferimenti per questa strategia?

%\begin{figure}
%    \centering
%    \includegraphics[width=\columnwidth]{./personal_space-2.jpg}
%    \caption{YYY.}
%    \label{fig:pers_space2}
%\end{figure}

\subsection{Person detection and pose estimation}
\label{sec:person_detection}
%\mc{qui metterei pedestrian detection and social distance estimation: una volta che ho la calibrazione e la posizione delle persone sul floor ho la social distance. La sezione successiva, mia e di Francesco, diventerebbe cosi' 'social distance characterization'}

Person detection has reached impressive performance in the last decade given the interest in automotive industry and other application fields \cite{benenson2014ten}. Real-time approaches can now estimate people pose even in complex scenario \cite{openpose:ijcv} and even reconstruct a 3D mesh of the person body \cite{guler2018densepose}. The majority of the approaches estimate not only people location as a bounding box but also 2D stick-like figures, so  conveying a schematic representation of the pose. Recently, several methods augment 2D poses in 3D or infer directly a 3D pose in a normalised reference system \cite{wang2014robust,moreno20173d,tome2017lifting,mehta2017vnect,martinez2017simple, bogo2016keep, mehta2019xnect,rogez2019lcr,zhou2017towards}.  

Capturing diffused small SDs with Computer Vision requires to individuate multiple people, realizing the hardest scenario for pedestrian detection techniques. Specific pedestrian techniques have been designed to work in crowded scenes~\cite{vandoni2019evidential,Wang_2018_CVPR,Liu_2019_CVPR,ge2020ps}, where skeleton-based representations are often drop in favour of saliency-based masks, especially focusing on heads. When the image resolution becomes too low to spot single people, regression-based approaches are employed~\cite{boominathan2016crowdnet,chan2008privacy,liu2018decidenet,setti2018count,wang2015deep,yang2003counting,padnet2020}, providing in some case density measures~\cite{sindagi2018survey,xu2016crowd,rangel2017entropy,sindagi2017cnn}. This information, merged with a geometric model of the scene, will directly lead to a measure of the average SD in the field of view. Obviously, regression or density-based approaches cannot provide additional cues on pose which are highly important for capturing human actions and interactions. To fill this gap, ad-hoc approaches individuate general crowd activities, classifying them as normal or not (e.g. a person collapsing and many people getting close)~\cite{direkoglu2017abnormal,gu2014abnormal,mohammadi2016groups,pennisi2016online}. %As an example, consider the case of a person collapsing and many people getting close to help him/her.
%Diffused small SDs would require a major intervention, like revisiting the architecture of the area, putting placeholders on the ground, etc., so imposing a different social behavior. 
%\vm{ATTENZIONE CHE SI STA DANDO UN SIGNIFICATO ELABORATO AD SD, CHE E' SOLO UNA DISTANZA}
%\mc{Non sto prendendoin considerazione solo la semplice distanza, ma il fatto che essa sia diffusa, ovvero che tutti la mantengano con qualcuno nella scena in analisi.}

Recently, new efforts tend towards solving the human detection and body pose estimation in crowded environments \cite{golda2019human,li2019crowdpose}, the very same scenario social distancing is dealing with.  %trying to prevent. 
Yet, finding the location of people in such cases is of relevant importance to alert or creating statistics of overcrowded areas. To this end, a people detection module has to be robust to  severe self and other objects/people occlusions, different image scales, and indoor/outdoor scenarios. Although a person detection (\ie, without the pose) may be enough for estimating the VSD, finding joints and body parts of pedestrian has certain advantages. This is due the fact that to obtain an approximate metric reference, or even calibrating cameras, usually the person height is used as a coarse proxy as computed from a bounding box or by more precise techniques \cite{wagner2019, vester2012estimating,benabdelkader2008statistical,dey2014estimating,Gunel_2019_ICCV}. However, bounding boxes do not account for different body poses (\eg, sitting, riding) that might negatively impact the estimate of height and thus a wrong VSD. Another issue is related to occlusions, \ie how much is reliable to extract a person height without having a full body information? This is necessary in the likely case of crowded environments or whenever an %occluding 
object partially hides person body parts (\eg, a person seated at a desk).

Given a metric reference from scene geometry and the position/pose of the people in the scene, the SD can be calculated as a distance on the ground plane (feet/body pose centroid) among all the possible detected pedestrians. As previously discussed, this information can be estimated locally or pairwise in order to reduce the complexity of estimating a global reference system for the whole image.

\subsection{Visual social distance characterization}
\label{sec:sd_chara}
Social distances should be complemented with additional contextual information to understand  whether social  distancing rules are actually being broken or not, suggesting as a consequence the most proper reaction.

\begin{figure}
    \centering
    \includegraphics[width=\columnwidth]{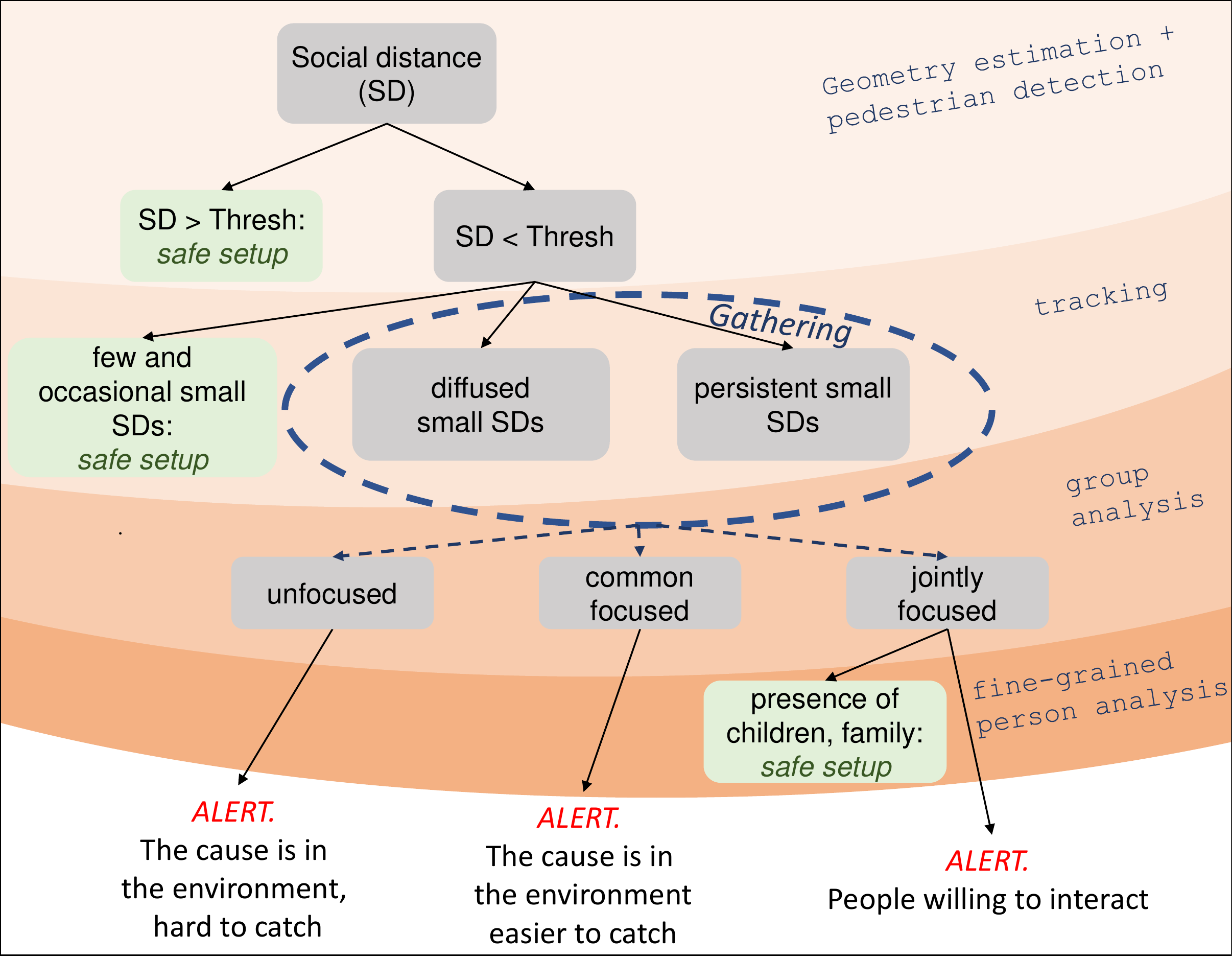}
    \caption{How to characterize social distances. Together with the taxonomy explained in the text, we specify in courier new the Computer Vision technologies to access a particular level of SD specification. The more detailed the SD characterization, the more advanced yet fragile Computer Vision technology is.}
    \label{fig:SDscheme_final}
\end{figure}

Fig.~\ref{fig:SDscheme_final} reports a multi-layer pipeline, which will be detailed in the following, indicating which information can be accessed with the current Computer Vision technology. The deeper the layer (indicated by a darker color), the finer the visual analysis which is needed and the harder the corresponding request for Computer Vision.
%%MARCO messo maiuscolo

As previously stated, SDs taking values above a certain threshold would certainly comply with social distancing rules.
%ensure the social distancing. \fs{da rivedere. Probabilmente intendi che i protocolli in materia di SD sono rispettati.}
%%MARCO: vero, rifrasato.
On the contrary, the presence of SDs under a certain threshold (\emph{SD$<$Thresh} in Fig.~\ref{fig:SDscheme_final}) \emph{could} be considered as breaking the rules, but actually many are the scenarios where this should not raise any concern. 

For example, occasional small SDs holding for few frames, especially in a crowded scenario (the \emph{few and occasional small SDs} blob in the figure), can be allowed, considering that they are detected by automatic approaches which are not exact.
%%MARCO IN figura, messo maiuscoli
Instead, small SDs can be critical if they are 1) \emph{diffused} and/or 2) \emph{persistent}.

In the former case, a high percentage of small SDs is characterizing the monitored area: This may occur at a crossing intersection or walking in a corridor of an airport. Here, many persons stand close, without an explicit will, possibly for a short ($\sim$seconds) period. Computer Vision helps here providing robust approaches for pedestrian detection and counting, discussed in Sec.~\ref{sec:person_detection}.

Persisting small SDs mean that \emph{specific people} stand close to each other \emph{for a certain time interval}. This condition addresses a more interesting situation, since it likely indicates people that stay close by intention. Under a Computer Vision point of view, persisting small SDs are more difficult to capture, since they require people to be tracked continuously, maintaining their identity. Interested readers may refer to~\cite{dendorfer2019cvpr19,liu2019estimating}, discussing the problem of tracking in crowded situations.

In Social Signal Processing, diffused and/or persisting small SDs individuate 
\emph{gatherings}~\cite{goffman1961encounters,goffman1966behavior,kendon1990conducting,setti2015f}, addressed generically as ``groups'' or ``crowds'' in Computer Vision.
The term gathering refers precisely to ``any set of two or more individuals in mutual presence at a given moment who are having some form of social interaction''~\cite{goffman1966behavior}. With the expression \emph{social interaction} we mean the process by which we act and react to those around us~\cite{setti2015f}.
Many types of gatherings are documented in the sociological literature, depending on: 
\begin{itemize}
    \item number of people being part of the gathering;
    \item \emph{type} of social interaction;
    \item spatial dynamics.
\end{itemize} 

As for the number of people, we may have small (2 to 6 people), medium (7 to 12-30 people), or large gatherings (larger than 13-31)~\cite{hare1981group}. 

Small gatherings happen in private (home, private garden, car), semi public (classroom, office, club, party area) and public places (open plaza, transportation station, walkway, park, street). 
Medium gatherings occur in private, but mostly in semi-public and public places, the latter being also the preferred venues of large gatherings~\cite{hare1981group}. 
%Large gatherings are most frequent in semi-public but mostly in public places~\cite{hare1981group}.
 
As for the type of social interaction, \emph{unfocused interaction} occurs whenever individuals find themselves by circumstance in the immediate presence of others. For instance, when forming a queue,
%\fs{questi esempi sono presi dalla letteratura? perchè a prima vista sembrerebbero common focused (mi aspetto che il focus di tutti quelli in fila sia il posto per cui stanno aspettando, così come la gente al semaforo dovrebbe avere come common focus il semaforo stesso).}
%%MARCO: Hai ragione Francesco, questi esempi sono strani. Me li ha dati Chiara bassetti. L'idea è che ad un semaforo io non ci vado perché voglio, ma perché mi tocca, quindi il semaforo non ha il mio interesse continuativo, ma solo occasionale, in special modo quando esso cambia. Infatti al semaforo tu vedi persone che si fa spesso i cavoli loro, mentre alle timetable la gente è li' per un interesse specifico. Stessa cosa per le code. Ho comunque tolto quello del semaforo, perché bastano due esempi. 
or when walking in the crowded corridor of an airport. On such occasions, simply by virtue of the reciprocal presence, some form of interpersonal communication must take place regardless of individual intent. 

For our study, having people forming an unfocused gathering and exhibiting small SD may indicate a problematic scenario, since it is the \emph{context} which encourages the formation of tight gatherings and not the will of people. As a consequence, to avoid such type of gathering may require a change of the context itself, for example discouraging the queues with markers on the floor, or creating lanes with barricades.  

Conversely, \emph{focused interaction} occurs whenever two or more individuals willingly agree -- although such an agreement is rarely verbalised -- to sustain for a period  a single focus of cognitive and visual attention~\cite{goffman1961encounters}.

Focused gatherings can be further distinguished in \emph{common focused} and \emph{jointly focused}~\cite{kendon1988goffman}.
In the former case, the focus of attention is common and not reciprocal, for example watching a timetable screen at the airport, watching a  map in the metro station, being at a concert. 
Common focused gathering exhibiting small SDs can be dealt more easily than in presence of unfocused gathering, since in this case the reason of the gathering is easier to be captured, which is the item or event attracting the common attention of people. 

Jointly focused gathering, finally, entails the sense of mutual, instead of merely common, activity. 
%A preferential openness to interpersonal communication, an openness one does not necessarily find among strangers at the theatre, for instance; in other words, a special communication license, like in a conversation, a board game, or a joint task carried on by a group of face-to-face interacting collaborators. 
%\vm{DA RIVEDERE E CHIARIRE}
%\mc{ho sistemato = tolto dalle balle le cose gia' espresse, che forse sono troppo "sociali"}
In a jointly focused gathering the participation, in other words, is not at all peripheral but engaged; people are -- and display to be -- mutually involved~\cite{goffman1966behavior}. 
%\mc{people in a jf are often in contact which each other}.
%
Since the presence of a jointly focused gathering depend on the will of people, when this is characterized by a small social distance, it can be discouraged by simply alerting the people about the ongoing critical setup. An exception for this scenario occurs when a jointly focused gathering involves children, since children have to be accompanied, and are usually at a physical contact with their relatives. 

\begin{figure*}
    \centering
    \includegraphics[width=\linewidth]{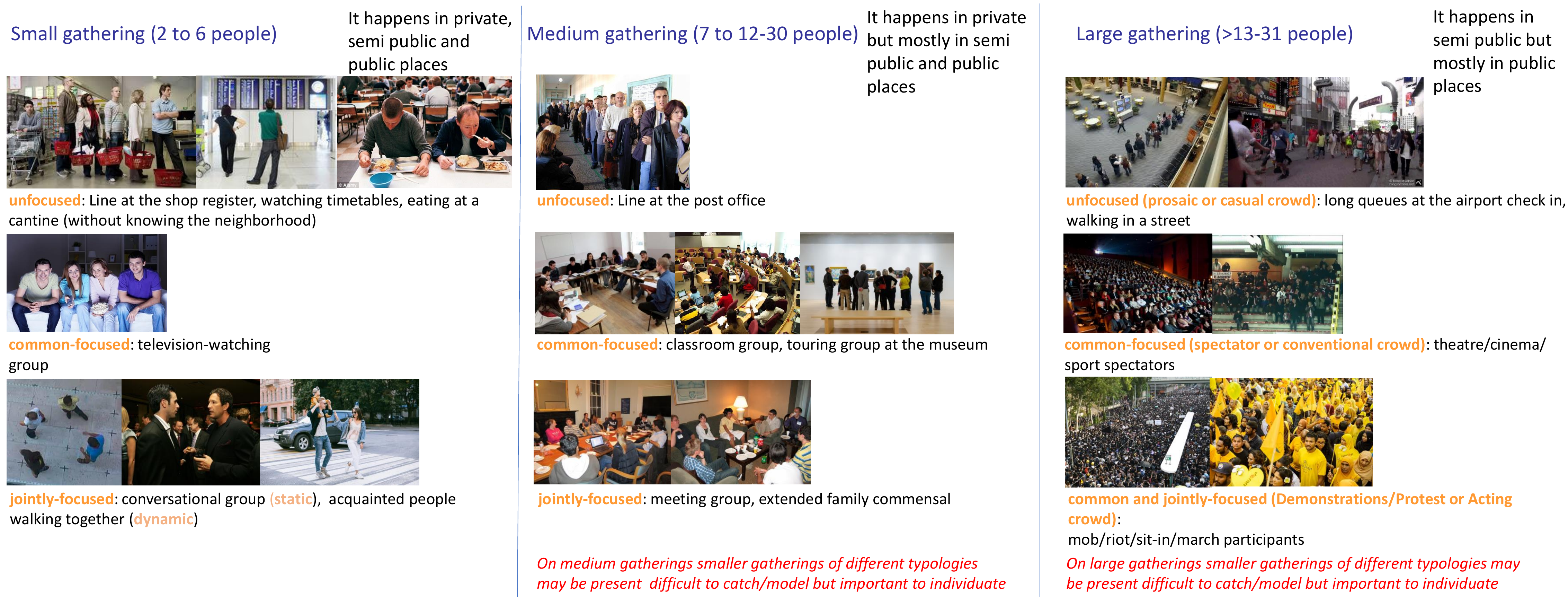}
    \caption{Different typologies of gathering, depending on the number of individuals involved and type of social interaction. (cfr. \url{https://vips.sci.univr.it/research/fformation/})}
    \label{fig:gatherings_type}
\end{figure*}

Some combinations of these attributes give rise to specific types of gatherings (shown in Fig.~\ref{fig:gatherings_type}), some of them addressed by explicit definitions: small gatherings of jointly focused people, mostly static, are dubbed by Kendon \emph{free-standing conversational groups}~\cite{kendon1990conducting}, highlighting their spontaneous aggregation/disgregation nature, implying that their members are jointly focused, and specifying their mainly-static proxemic layout. Large gatherings of unfocused people are named \emph{casual crowds}~\cite{blumer1957collective}; commonly focused large gathering refers to \emph{spectator crowd}~\cite{blumer1957collective} and, finally, large gatherings of jointly focused people are \emph{demonstration/protest} or \emph{Acting crowd}~\cite{mcphail1983individual}.

As anticipated above, most Computer Vision approaches do not build upon this taxonomy, distinguishing merely gatherings depending on the number of individuals involved, leading to groups (= small gathering for sociology) and crowds (= large gatherings), with some exception presented in the following.
Groups have been usually identified exploiting positional and velocity cues (people in a group is close and move with similar oriented velocity)~\cite{groh2010detecting_3,rota2012eccv,hung2011detecting,setti2013wiamis,solera2016pami,swofford2018cnn,swofford2019dante,sanghvi2019mgpi,tran2013social}. 
Explicit focus on free-standing conversation groups is given in~\cite{Cristani:FF:BMVC:2011,inaba2016conversational,setti2015f,setti2013icip,vascon2016cviu,vascon2019}. In most of these latter approaches, positional and velocity cues are enriched by pose information, fully capturing the people proxemics.
Coming back to the characterization of SDs, and to Fig.~\ref{fig:SDscheme_final}, joint-focused groups where people stand closer than a given threshold requires maximal attention, since their vicinity is by choice, and not by external circumstances. At this level of characterization, avoiding false alarms would mean to focus on the age of the interactants: Having children in a small gathering would probably indicate a family. Approaches like\cite{yuan2018does} estimate the age from pedestrian detections, but solving this task efficiently seems to be still at its early stages. 

%Traditional approaches rely on proxemics cues to infer pairwise interactions~\cite{groh2010detecting_3,rota2012eccv} or to identify clusters~\cite{hung2011detecting,inaba2016conversational,setti2015plos,tran2013social,Cristani:FF:BMVC:2011,setti2013wiamis,setti2013icip,vascon2016cviu,vascon2019}. More recently, deep learning solutions have been proposed~\cite{sanghvi2019mgpi,swofford2018cnn,zhuang2017drcnn,swofford2019dante,fernando2018cdgan}.

As for the modelling of crowd, some approaches allow to estimate the number of individuals~\cite{boominathan2016crowdnet,chan2008privacy,liu2018decidenet,setti2018count,wang2015deep} or their density~\cite{sindagi2017cnn,sindagi2018survey,xu2016crowd}.
Social Signal Processing approaches for large gatherings focus specifically on common-focused formations (aka \emph{spectator crowd}), here too capturing proxemic cues including the body pose~\cite{conigliaro2013viewing,conigliaro2015shock}.
Medium gatherings have never been properly addressed neither by Computer Vision nor Social Signal Processing literature.

%\mc{questo qui sotto lo vorrei depotenziare come messaggio}
%\fs{Io lo rimuoverei del tutto, non aggiunge molto a quanto detto fino ad ora.}

Finally, we should consider the static/dynamic axis concerning the degree of freedom and flexibility of the spatial, positional, and orientational organisation of gatherings. Distinguishing between uncommon, common-focused and jointly focused is hard, since when a gathering is moving, their members spend attention to follow safe trajectories, avoiding collisions. Therefore, the aforementioned taxonomy holds especially for static formations, with few exceptions~\cite{ciolek1980environment}. When people are moving, the only valid distinction that Computer Vision and Social Signal Processing do follow is that of small and large gathering.  

For small gathering, temporal information allows one to provide stronger grouping estimations, analyzing pedestrians motion paths instead of static positions~\cite{mazzon2013detection,solera2016pami,voon2019collective}.
For large gathering, many approaches identify dominant flows and segment crowd according to coherent motion~\cite{cheriyadat2008motions,kang2014fully,tu2008unified,wang2014finding}, and to identity collective/abnormal behaviors~\cite{direkoglu2017abnormal,gu2014abnormal,mohammadi2016groups,pennisi2016online}.

Summarizing, once small SDs are detected, it is necessary to understand if they are persistent and/or diffused in the scene. Then, proxemic analysis is needed to characterize the gatherings which are generating the SDs. Unfocused gatherings would indicate SDs are caused by no explicit will; common-focused gatherings come usually because of the presence of precise environmental conditions (a manufact or an event attracting the attention); jointly-focused gatherings indicate explicit will of interacting, and could be further described by capturing the age of interactants (kinship). Each of these formations may demand for different interventions, thus going beyond the simple alarm when SDs are too small, and diminishing false positive alarms.

Computer vision approaches following this taxonomy exist for jointly focused (small) gatherings, (large) common focused gatherings, and show that positional and body pose cues are of primary importance. Future work has to be done to cover all the possible types of gatherings, as current technology is still struggling to achieve a solution, especially when they are composed by several people.

\section{Beyond social distancing: Applications}
%
% Ho messo alcune arre di cui sono a conoscenza, il testo puo' essere ulteriormente ampliato aggiungendo
% piu' paragrafi.
%

%%% An intro for the redesign of architecture.
% A clear impact of the virus outbreak is related to redesign of our spaces for allowing to restart our daily life and possibly respecting social distancing... 

While potentially playing a crucial role in the case of a virus outbreak, technology developed for the analysis of social distancing can be useful in a large number of application domains that, therefore, can benefit from the approaches proposed in this work. 

The detection of mental health issues is one of the areas that will benefit most from the application of AI\footnote{According to the \emph{Gartner Group}, a relevant strategic consulting company: \url{http://www.gartner.com/smarterwithgartner/} \url{13-surprising-uses-for-emotion-ai-technology/}}, supported by the World Health Organization observing that a pathology like depression affects around 300 millions people around the world~\cite{who2017}. In such a particular case, the tendency to avoid physical proximity and engagement with others is an important symptom. The technologies proposed in this work can also help the increase of SD, especially when it is hard to observe. Similarly, the analysis of interpersonal distances can help to identify children with insecure attachment, known to manifest their condition through irregular proximity patterns (among other cues)~\cite{Bowlby1969}.

Another important domain  where the analysis of SD is important is social robotics. In particular, the \emph{International Federation of Robotics} pointed out that public relation robots are the fastest growing area of service robotics with estimates in sales moved from a total of USD 319 million in the period 2015-2017, to a total of USD 746 million between 2018 and 2020.%\footnote{\url{https://ifr.org/downloads/press/Presentation_PC_11_Oct_2017_1.pdf}}. 
In this field, %The literature shows that people tend to interpret the behaviour of robots, whether they have human-like appearance or not, in the same way as they interpret the behaviour of people, an effect called \emph{media equation}~\cite{Reeves1996}. In particular, 
the use of proxemics appears to be particularly important to ensure that a robot is perceived to play correctly its role (\eg whether it is expected to be a servant or a companion in playing)~\cite{Koay2014} and to establish a sense of intimacy~\cite{Kennedy2017}, an aspect of focal importance in assistive robotics. In addition, distance plays a major role along one of the five Godspeed dimensions typically used to assess the quality of human-robot interaction, namely perceived safety~\cite{Bartneck2009}. 

%\vm{1) NON SO SE VAL LA PENA QUI PARLARE ANCHE CHE L'INTERAZIONE/DISTANZA human-robot VA PARTICOLARMENTE CURATA PER LA SICUREZZA DELLA PERSONA. 2) OLTRE A CIO' VOGLIAMO DIRE QUALCOSA ANCHE SUI COBOT - COLLABORATIVE ROBOT - IN AMBIENTE LAVORATIVO?}

In the last years, most major companies have introduced training to avoid unconscious bias, \ie the tendency to discriminate certain categories of people without being aware of it. This happens not only for ethical reasons, but also because \emph{McKinsey} %, a leading strategic management firm, 
has shown that companies ensuring diversity in their workforce, especially at the top management levels, are 30\% more likely to be above national median in terms of financial returns~\cite{Hunt2015}. As a consequence, major companies like Facebook (\url{https://managingbias.fb.com}) and Google (\url{https://diversity.google}) adopt implicit training programs. Furthermore, Forbes estimates that the market of implicit bias and diversity training has reached a value close to USD 9 billion-a-year (\url{http://goo.gl/R53xn4}). Unconscious bias leaves different traces in nonverbal behaviour and one of these is the increase of physical distances (see, \eg~\cite{McCall2009}). Therefore, automatic technologies for proxemics analysis can  help to detect the phenomenon, contributing to protect the potential victims, and train the bias bearers to identify and attenuate their tendencies to discriminate others. 

A large number of studies show that the architectural design of space influences the behaviour of its inhabitants~\cite{Mallgrave2013}. For example, a simple line on the floor separating right and left side of a corridor makes the flow of people through it more ordered~\cite{Ball2004}. Similarly, the restructuring of Westminster in the UK aims at improving the efficiency of parliamentary works, but encounters the opposition of Parliament workers afraid of disrupting established traditions by the change 
of the way space is organised~\cite{Siebert2019}.
Until now, the study of these phenomena has been performed mainly through ethnographic observations, but the development of technologies for proxemic analysis can certainly help by producing more objective and quantitative data about the change in habits of the people. This is in line with previous works about the study of organisations through the use of smart badges detecting who is in proximity with whom in an organisation~\cite{Eagle2014}.

Besides the application scenarios above, likely to benefit from the technologies presented in this work in the future, there are established domains that can benefit from models of mutual distancing. For example, Augmented and Mixed Reality technologies can provide more immersive and engaging experiences through the inclusion of virtual characters capable to move with respect to users like humans do with respect to one another. Similarly, surveillance systems can further refine their ability to detect events of interest in a given environment like, e.g., an aggression in a public space. Finally, technologies analysing interpersonal distances can be of help to social psychologists that investigate the dynamics of social interactions. In other words, far from being exhaustive, the list of  application domains listed in this section still provides an indication of how wide  the application of VSD can be once the Covid 19 outbreak, at the origin of the most recent interest towards interpersonal distances, will be over.

%\mc{\textbf{[All]}: everybody adds future and futuristic scenarios where social distance has to be considered: design of buildings, theme parks. conferences... How virtual place can transmit what is trasmitting the social distance: virtual rooms? symbolic approaches?  }
%\\
%\vm{Just rough ideas in sparse order of what we can write, to be better elaborated:
%\begin{itemize}
%\item AR/VR: mix of avatar or hologram representations in special setup (room), usage of visors or glasses or no need, need of special (rendering, visualization) technology to focus at close range; 
%\item new buildings, taking into account special walking path and sitting areas;
%\item new public spaces, indoor and outdoor (even footsteps): same as before;
%\item shops, commercial centers and shopping activities;
%\item every public service (banks, post offices, etc) should comply to ...;
%\item working places: offices -> easy, assembly and production line -> revision, ... others?;
%\item big events: concerts, sports, parades
%\item kindergarten, schools and universities: these will likely experience a major issue, given the limited space available;
%\item Public transports: buses, trains, airplanes: even big changes
%\item ...???
%\end{itemize}
%
%}

\section{Privacy and acceptability concerns}
%\mc{\textbf{[Alessandro + Vittorio]}}
%\vm{written something}
%\\

Optical cameras are the most widespread sensors for VSD measurements and the acceptance of this monitoring technology can be difficult since it clearly raises privacy concerns. Video footage may disclose the identity of the persons captured and in general recording is regulated by strict laws, both at national and international level. Moreover, potential attacks to the video transmission channels and to storage servers can pose a relevant security issue. %As a whole, this may impose restrictions on the utilization of such visual sensors and raise acceptability concerns from the people.
%, leaning the usage of more "anonymous" devices. 

However, the current computer vision technology is now mature to manage effectively privacy concerns. %, in two different ways at least. Both
Alternatives benefit from the usage of the so-called \textit{smart} cameras~\cite{belbachir2010smart}, which have computing capability onboard, able to process video data up to a certain capacity. By adopting a privacy-by-design principle, a first option is to process video sequences internally, while measuring and transfer only VSD estimates without any image, thus sensible data, being transmitted to the remote control operative room. This is of crucial importance for VSD, since as we have been shown in the previous sections, accurate estimates may requires the identification of kinship. This sensible information clearly is not necessary to disclose for estimating VSD and any possible leak has to be avoided.

%A second option can operate by executing persons' detection onboard and substituting each detection by an anonymous, synthetic human figure (like a stick figure, see Fig.~\ref{???}), or blurring faces (or the entire body), before transmission, so to make impossible to recognize the identity. In all cases, any transmission and storage of the acquired raw video data is avoided, so preventing privacy violations and security concerns. 

At the same time it is worth noting that VSD technology exhibits features which differentiate it from other apparently safer alternatives, as geolocation data collected from mobile applications. VSD techniques are in fact non-invasive and mostly non-collaborative, meaning that the user does not need to provide ID personal data. Tracing technologies, on the contrary, need to be fed with sensible data and even when this is totally anonymized, recent research~\cite{de2018privacy} proves that individuals may still be identified by a few information – four spatio-temporal points allows one to uniquely identify 95\% of people in a mobile phone database of 1.5 million subjects and 90\% of people in a credit card database of 1 million individuals.

\section{Conclusions}
%\mc{\textbf{[Alessio, Marco]}}
In this paper we have presented the VSD problem as the estimation and characterization of inter-personal distances from images. Solving such problems allows a quick screening of the population for detecting  potential behaviours that can cause a health risk, especially related to recent pandemic outbreaks. We pointed out that VSD is not only a Computer Vision problem related to geometrical proxemic since people distancing has to be weighted given the social context in the current scene. Close relationships can allow closer interpersonal distances as well as being a caretaker of individuals with fragile conditions. We have shown that understanding such social context is a compelling problem in the literature of signal social processing that requires further research efforts for a reliable solution. As the solution is intertwined with the decoding of social relationship from images, there are strong ethical and privacy concerns that need to be addressed with novel privacy-by-design solutions. Past this grievous global crisis, VSD has still an important role in several application fields thus providing a continuous source of interest in this new problem. 

\section*{Acknowledgment}

The authors would like to thank P. Morerio, M. Gavari, G.L. Bailo for the results on social distancing estimation in the figures.
This work has been partially supported by the projects of the Italian Ministry of Education, Universities and Research (MIUR) ``Dipartimenti di Eccellenza 2018-2022''.

\bibliographystyle{ieee}
\bibliography{egbib,mybib}


%\end{IEEEbiography}


%\newpage



%\vfill


\end{document}